# Easy-to-Read in Germany: A Survey on its Current State and Available Resources

**Margot Madina[1], Itziar Gonzalez-Dios[2], Melanie Siegel[1]**

[1]Darmstadt University of Applied Sciences (Hochschule Darmstadt)
margot.madina-gonzalez@h-da.de
melanie.siegel@h-da.de
[2]HiTZ Zentroa-Ixa, Euskal Herriko Unibertsitatea (UPV/EHU)
itziar.gonzalezd@ehu.eus

**Abstract**
Easy-to-Read Language (E2R) is a controlled language variant that makes any written text more accessible through the use of clear, direct and simple language. It is mainly aimed at people with cognitive or intellectual disabilities, among other target users. Plain Language (PL), on the other hand, is a variant of a given language, which aims to promote the use of simple language to communicate information. German counts with Leichte Sprache (LS), its version of E2R, and Einfache Sprache (ES), its version of PL. In recent years, important developments have been conducted in the field of LS. This paper offers an updated overview of the existing Natural Language Processing (NLP) tools and resources for LS. Besides, it also aims to set out the situation with regard to LS and ES in Germany.

**Keywords:** Easy-to-Read, Leichte Sprache, einfache Sprache, readability, accessibility

## 1. Introduction

Access to information, knowledge and culture is a right for all citizens. As stated in the Convention on the Rights of Persons with Disabilities of the United Nations (CRPD), access to information and communication is a fundamental right, and states should facilitate information in accessible ways such as easy to read and understand forms (United Nations, 2006). However, written text does not always match our ability to understand what we read; a lot of the textual information we find nowadays is too complicated to understand by people with communication disabilities. This causes their exclusion from society. Easy-to-Read (E2R) is a pivotal part of the inclusion for people with communication disabilities (Hansen-Schirra et al., 2020; Maaß and Hansen-Schirra, 2022).

E2R is a language variant of a standard language, with reduced complexity, and with the aim to improve the readability and comprehensibility of texts (Nitzke et al., 2022). One of its functions is to make content accessible, and to ensure participation for people with communication impairments (Hansen-Schirra and Maaß, 2020). It counts with a set of rules that cover the vocabulary, grammar structures and layout of the text, among others.

E2R is mainly aimed towards people with cognitive or intellectual disabilities; however, other target groups may also benefit from it. E2R's target groups include, but are not limited to, the following: people with intellectual, cognitive or developmental disabilities, people with auditory disabilities, people with low literacy, migrants, or children in need of reading reinforcement (Bredel and Maaß, 2016; Hansen-Schirra and Maaß, 2020; Maaß and Garrido, 2020; Maaß, 2019).

E2R usually receives a different name depending on the standard language it is based on; in the case of German, its E2R variant is known as *Leichte Sprache* (LS). It is not to be confused with what is known as Plain Language (PL) *Einfache Sprache* (ES) in German). PL is a variant of a given language, which aims to promote the use of simple language to communicate information. There are some important differences between them: (1) PL focuses on text while E2R covers text, illustrations and layout, (2) E2R is usually aimed at people with intellectual disabilities, as PL might be too challenging for them, (3) PL was initially focused on legal and governmental texts, due to their intrinsic meaning, while E2R is usually applied to all sorts of texts.

The rest of the paper is structured as follows: Section 2 will discuss the current state of LS in Germany. Section 3 will overview the existing resources and tools for LS, briefly describing how they are relevant in the LS field. Finally, Section 4 will present the conclusions.

## 2. LS in Germany

The first rules for LS were developed by Inclusion Europe back in 1998 (Pottmann, 2019). However, this first set of rules, even thought they were written in German, were generic rules for E2R that could be applicable to any standard language. In 2002, the German government was obliged to provide accessible information to everyone due to the establishment of the *Gesetz zur Gleichstellung von Menschen mit Behinderungen* (the German equality law for disabled people) and the Barrierefreie-Informationstechnik-Verordung (the accessible technology enactment). The *Netzwerk Leichte Sprache* (the plain language association) was founded in 2006, and they developed the rules for LS in 2013 (Pottmann, 2019)[1].

There is currently an ongoing debate related to LS rules, terminology, and its possible stigmatization effect. At present, there are three main currents regarding LS in Germany.

Andreas Baumert defends the use of ES over LS, claiming LS to be "falsches Deutsch" (bad German) (Baumert, 2018, 3) that cannot be used by many of its target groups. He strongly criticizes LS and, among the shortcomings he found, he says that it is all a business and that it cannot be generalized for all target groups. In his view, ES has a better chance of taking shape in the foreseeable future; however, there is no generally accepted version of a simple language that is used by authorities, industry and associations alike (Baumert, 2016, 94).

---
[1] For a more detailed overview of how the legal state of LS has changed over time, refer to Maaß, 2020

Bettina Bock proposes that LS should not be understood as strictly bound to rules; even though they might be useful, they should not be seen as strict norms but rather as suggestions. She claims that LS should be used as an umbrella term that covers all approaches that this term encompasses. The focus should be on the users of LS and seek what is best for them, not so much on the rules, as there are some texts in LS that, even though they do not adhere strictly to the rules, are considered good by the community (Bock, 2018, 11).

Christiane Maaß advocates for the use of LS and the implementation of its rules. She proposed a set of rules for LS (Maaß, 2015) and is currently working on making them more precise and their regularization. She also claims that the term "Easy-to-Read" is not an adequate term for this language variant, as it is used in other forms of realization such as screen readers, sightseeing or museums, or interpreting in inclusive meetings and conferences. She proposes the term "Easy Language" (EL) instead, as it is open to broader conceptualizations (Maaß, 2020, 56). Besides, she also highlights that E2R and LS texts may have a stigmatizing effect on their target groups. Therefore, she proposes a new model, Leichte Sprache Plus (Easy Language Plus), which stands between PL and E2R (in the case of German, between LS and ES). This model "profits from the comprehension and perception principles of EL, but also from the non-stigmatising and more acceptable features of PL" (Hansen-Schirra and Maaß, 2020, 32) However, it is not an established approach yet, but it is only on its first steps.

Nowadays, LS has a very active practice in Germany, while ES is rarely used. A reason for this might be that it is safer for public bodies to go for LS, as ES might still be challenging for some target users, whereas LS reaches a wider audience (Maaß, 2020).

## 3. Resources and Tools for LS

This section will introduce the available resources and tools for LS. We concentrate on the most recent and relevant ones related to LS.

### 3.1. LS Corpora

A corpus (plural, corpora) is a linguistic resource consisting of a large, structured set of texts. It can be parallel (combines a simplified version of a text with its original version) or comparable (a collection of simplified documents and standard-language documents that share the same topic).

**Klaper et al., 2013**: Klaper et al. developed the first parallel, sentence-aligned corpus with German and LS texts. They crawled the data from five publicly available webpages. The corpus consists of around 70,000 tokens, and spans various topics. The quality of the sentence alignment obtained an F-score of 0.085.

**LeiSa** (Lange, 2018): this corpus was created in order to do an explorative corpus-based analysis as part of the research project *Leichte Sprache im Arbeitsleben* (LeiSA, Easy-to-Read in Work Contexts, University of Leipzig). It is a collection of texts, but it is not a parallel nor aligned corpus. The aim was to obtain a systematic corpus-based description of the distinctive linguistic structures of LS by contrasting it to similar approaches of text simplification (i.e. *einfache Sprache* and *Leicht Lesen*). The corpus contains 639,826 tokens of LS, 779,278 tokens of *Einfache Sprache*, and 350,872 tokens of *Leicht Lesen*.

**Battisti and Ebling, 2019**: this corpus was compiled to be used in automatic readability assessment and automatic text simplification (ATS) in German. It was compiled from web sources and consisted of both monolingual (LS) and parallel data (German and LS). It also contained information on text structure, typography and images; according to their authors, these features can indicate whether a text is simple or complex. The monolingual data consists of 1,916,045 tokens and the parallel data consists of 347,941 tokens of German and 246,405 tokens of LS. However, there was no sentence-alignment in this corpus.

**TextComplexityDE** (Naderi et al., 2019a): this dataset consists of 1000 sentences taken from 23 Wikipedia articles in 3 different article-genres. 250 of those sentences have also been manually simplified by native speakers. Besides, it also contains subjective assessment (complexity, understandability and lexical difficulty) of the simplified sentences, provided by a group of language learners of A and B levels. This dataset is aimed to be used for developing text-complexity predictor models and ATS.

**APA** (Säuberli et al., 2020): the Austria Presse Agentur (APA) corpus is the first parallel corpus for data driven ATS for German. It consists of 3,616 sentence pairs. The authors manually simplified original sentences into their A2 and B1 equivalents and aligned them.

**LeiKo** (Jablotschkin and Zinsmeister, 2020): it is a comparable corpus of LS news texts, systematically compiled and linguistically annotated for linguistic and computational linguistic research. It contains approximately 50,000 tokens, and is divided into four sub-corpora according to the websites from which they were extracted.

**KED** (Jach, 2020): Korpus Einfaches Deutsch (KED) is a collection of texts from genres of educational and public discourse in LS and *Einfache Sprache*, scraped from different online websites. It has a total of 3,698,372 words, and it is divided into different sub-corpora depending on the provider.

**20m** (Rios et al., 2021): it is a corpus collected from the Swiss news portal *20 Minuten*, which includes 18,305 articles paired with shortened summaries. There is no sentence-alignment in this corpus, and the dataset does not distinguish different simplification levels; they do not stick to any simplification standard.

**Capito**: *capito[2]* is the largest provider of human simplification services for German; they translate information into easy-to-understand language, offer trainings, and develop digital solutions around the topic of comprehensibility. It has a dataset that covers a wide range of topics and levels A1, A2 and B1 (Rios et al., 2021).

**Geasy** (Hansen-Schirra et al., 2021): the German Easy Language corpus (Geasy) is a parallel corpus, aligned at sentence level, which contains professional translations from standard German into LS. It currently contains 1,087,643 words of source text and 292,552 words of LS translations.

---

[2] https://www.capito.eu (last accessed: 2022-11-04)

**Toborek et al., 2022**: this is a new monolingual sentence-aligned corpus for German, LS and ES, spanning different topics. This corpus consists of publicly available articles of 7 different webpages that publish news articles in German and their corresponding LS version. They also included articles from a website in ES in an aim to achieve a larger vocabulary size. They refer to all simplified versions of German as *Simplified German*. The corpus has a total of 250,093 tokens of *Simplified German* and 404,771 tokens of German, contains 708 aligned documents and a total of 5,942 aligned sentences. The quality of the sentence alignments has a F1-score of 0.28.

**SNIML** (Hauser et al., 2022): simple news in many languages (SNIML) is a multilingual corpus of news in simplified language. It includes articles in Finnish, French, Italian, Swedish, English and German, published between 2003 and 2022, and originates from different news providers in different countries. It is a dataset of raw text. Besides, the level of simplification varies depending on the provider, that is, the texts have been created according to different simplification guidelines and for different target audiences. However, the authors claim that the corpus is useful for automatic readability assessment and for unsupervised, self-supervised or cross-lingual learning. They plan to release a new version of SNIML every month, and their future work may consist of aligning the articles to related articles in standard language. By the time this article is being written, it contains 4,936,181 tokens in total, 123,021 of which are of German.

**Klexicon** (Aumiller and Gertz, 2022): this is a document-aligned corpus by using the German children encyclopedia "Klexikon". It contains 2,898 articles from "Klexicon", with an average of 436.87 tokens each, and 2,898 documents from Wikipedia, with an average of 5,442.83 tokens each. The authors aligned the documents by choosing corresponding articles from Wikipedia; however, it is unlikely that specific sentences are matched.

### 3.2. Other Resources

#### 3.2.1. Dictionaries

E2R texts may also include explanations of certain terms. E2R dictionaries can be useful for this end, as they provide translations from a standard language term into its E2R equivalent, or an explanation for complex words that cannot be adapted into an easier variant. German counts with *Hurraki*[3], a dictionary with explanations of German words in LS.

#### 3.2.2. LS Language Checkers

They are employed to check texts for grammar and style mistakes. For E2R, they can help checking whether any E2R rule has been broken. Siegel and Lieske (Lieske and Siegel, 2014; Siegel and Lieske, 2015) implemented some EL rules in Acrolinx[4] and LanguageTool[5].

### 3.3. Tools for LS Adaptation

---
[3] https://hurraki.de/wiki/Hauptseite (last accessed: 2022-11-03)
[4] Acrolinx is a software package to support authors of technical documentation
[5] LanguageTool is an open source text checking software developed since 2003

E2R adaptation refers to the processes that standard texts undergo in order to be transformed into E2R texts. These processes may include syntactic simplification (reducing the grammatical complexity of a text), lexical simplification (replacing complex words with easier variants) or summarization (conveying the most important information), among others.

**EasyTalk** (Steinmetz and Harbusch, 2020): EasyTalk is a system for assisted typing in LS. It uses a paraphrase generator based on a lexicalized, unification-based Performance Grammar.

**SUMM**[6]: this is the first AI-powered tool that automatically turns any text into EL. It is still only available in its Beta version, but its founders claim that it increases adaptation productivity by 85%. Once registered, users can directly go to the translation interface and adapt any text. Currently, the glossary is based on *Hurraki*, although users can also create their own glossary based on their texts.

### 3.4. LS Readability Assessment

Readability assessment is used to classify texts according to their degree of complexity; it determines how difficult or easy a text is or which level/grade it has (Bengoetxea and Gonzalez-Dios, 2021; Vajjala, 2022). It can help authors prepare simplified material, inform readers about the difficulty of a piece of text, or facilitate choosing of learning material for second language learners, among others (Aluisio et al., 2010).

**Battisti et al., 2019**: they present an unsupervised machine learning approach to analyse texts in simplified German in an aim to investigate evidence of multiple complexity levels. They also exploited structural and typographic characteristics of simplified texts. Their findings prove that there is not just one complexity level in German simplified texts.

**Ebling et al., 2022**: this is the first sentence-based NMT approach towards automatic simplification of German and the first multi-level simplification approach for German. Besides, this paper offers an overview of four parallel corpora of standard/simplified German, compiled and curated by their group. They report a gold standard of sentence alignments from these four sources.

**Naderi et al., 2019b**: in this study, they developed an automated readability assessment estimator based on supervised learning algorithms over German text corpora. They employed the TextComplexityDE corpus. They extracted 73 linguistic features and employed feature engineering approaches to select the most informative ones. They implemented 4 regression estimators to assess the readability of the sentences, among which Random Forest obtained the best result, with a 0,847 RMSE.

**Mohtaj et al., 2022b**: they present a new model for text complexity assessment for German text based on transfer learning. They used the TextComplexityDE dataset to train the models. Their findings show that fine-tuning the BERT model can outperform the other approaches.

**Weiss and Meurers, 2022**: this study presents a sentence-wise readability assessment model for German L2 readers. They built a machine learning model with linguistic insights and compare its performance based on predictive regression and sentence pair ranking. They found that it yielded top performances across tasks.

---
[6] https://summ-ai.com/en/ (last accessed: 2022-11- 03)

**Blaneck et al., 2022**: in this study, they combined the fine-tuned GBERT and GPT-2-Wechsel models with linguistic features. They evaluated their models in the GermEval 2022 Shared Task on Text Complexity Assessment with the TextComplexityDE dataset. The combined models performed better than non-combined GBERT or GPT-2-Wechsel models. On out-of-sample data, their best ensemble achieved a RMSE of 0.435.

**Mosquera, 2022**: this paper describes the winning approach in the first automated German text complexity assessment shared task as part of KONVENS 2022. The only resource provided by the organizers was the TextComplexityDE dataset. They followed two main approaches to train the dataset: feature engineering based on morphological and lexical information, and transfer learning via pre-trained transformers.

**Mohtaj et al., 2022a**: this paper offers an overview of the GermEval 2022 Shared Task on Text Complexity Assessment of German Text. Due to space constraints, not all studies that took part in it have been included in the paper at hand. However, this study offers an overview of the task and the approaches. Among 24 participants who registered for the shared task, ten teams submitted their results on the test data.

## 4. Conclusion

This paper presents the current situation of LS in Germany and offers an overview of the most important developments regarding resources and tools for LS adaptation. It can be stated that there is a dynamically developing research situation of LS in Germany. Nonetheless, as it has been highlighted, there are also different currents with regard to LS rules, terminology and its possible stigmatising effect. In spite of LS being the established language variant, there is a lack of consensus regarding the accuracy of its name and the users it is aimed at. This may lead to potentially remodeling LS rules and validating *Leichte Sprache Plus* in the near future (Lindholm and Vanhatalo, 2021, 209). It is possible that the concept of LS and E2R will be reshaped, and different names will be used to refer to them; however, to this day, it would be advisable to stick to the terms "Leichte Sprache" and "Easy-to-Read", since these are the established terms and the use of other terminology might cause confusion. Regarding the available resources and tools, it is worth highlighting that there is a recent interest in developing corpora. Nonetheless, it can be observed that there is no consistency within the databases; they might be parallel, comparable, aligned, and can include texts of different complexity levels. This might be due to different reasons: (1) they have been thought to serve for different purposes, or (2) there is not enough data to create the corpora from. Some webpages offer information both in standard German and E2R, but some others do not. This makes it difficult to create parallel, sentence-aligned corpora. Besides, many so-called E2R resources that can be found in many websites are often not linked to a single corresponding German document, but are high-level summaries of multiple German documents (Klaper et al., 2013). Many readability assessment methods have been developed recently; it would be helpful for LS and ES users to see these methods implemented, so that they can assess the difficulty of a given text. The existing LS adaptation tools may aid in the creation of LS texts; however, they are still not able to automatize the insertion of examples, explanations and illustrations or to create a proper layout for LS. Taking everything that has been said into account, we could say that E2R and ATS have been a recurrent field of study in these recent years and seem to stay that way. However, much remains to be done, especially in terms of developing products based on all available tools and resources. These products would make information more accessible to people with communication disabilities.